\documentclass[journal,twoside,web]{ieeecolor}
\usepackage{etoolbox}
\makeatletter
\@ifundefined{color@begingroup}%
{\let\color@begingroup\relax
\let\color@endgroup\relax}{}%
\def\fix@ieeecolor@hbox#1{%
\hbox{\color@begingroup#1\color@endgroup}}
\patchcmd\@makecaption{\hbox}{\fix@ieeecolor@hbox}{}{\FAILED}
\patchcmd\@makecaption{\hbox}{\fix@ieeecolor@hbox}{}{\FAILED}

\usepackage{tmi}
\usepackage{orcidlink}
\usepackage{tikz}
\usepackage{cite}
\usepackage{amsmath,amssymb,amsfonts}
\usepackage{bbm}
\usepackage{hyperref}
\hypersetup{hidelinks}
\usepackage{algorithmic}
\usepackage{graphicx}
\usepackage{textcomp}
\usepackage{booktabs} 
\usepackage{makecell}
\usepackage{pifont}

\usepackage{pifont}
\newcommand{\cmark}{\ding{51}}%
\newcommand{\xmark}{\ding{55}}%
\usepackage{bbold}

\usepackage{stfloats}
\usepackage{subfigure}
\usepackage{colortbl}  
\usepackage{array}  

\usepackage{multirow} 
\usepackage{colortbl}

\definecolor{myblue}{RGB}{212, 239, 251}
\definecolor{mygreen}{RGB}{217, 255, 217}
\definecolor{myyellow}{RGB}{255, 252, 200}

\definecolor{deepblue}{cmyk}{0, 0.5, 0.5, 0}

\def\BibTeX{{\rm B\kern-.05em{\sc i\kern-.025em b}\kern-.08em
    T\kern-.1667em\lower.7ex\hbox{E}\kern-.125emX}}
\begin{document}
\title{Self-Paced Learning for Images of Antinuclear Antibodies}

\author{Yiyang Jiang\hspace{-1.2mm}$^{~\orcidlink{0009-0007-1169-4465}}$, \IEEEmembership{Student Member, IEEE}, Guangwu Qian\hspace{-1.2mm}$^{~\orcidlink{0000-0001-9241-1699}}$, Jiaxin Wu\hspace{-1.2mm}$^{~\orcidlink{0000-0003-4074-3442}}$, Qi Huang, Qing Li\hspace{-1.2mm}$^{~\orcidlink{0000-0003-3370-471X
}}$, \IEEEmembership{Fellow, IEEE}, \\[-1.5mm]
Yongkang Wu and Xiao-Yong Wei\hspace{-1.2mm}$^{~\orcidlink{0000-0002-5706-5177
}}$, \IEEEmembership{Senior Member, IEEE}
\thanks{This research was supported by the Hong Kong Research Grants Council via the General Research Fund (project no. PolyU 15200023) and the National Natural Science Foundation of China (Grant No.: 62372314). The experimental part of this work was supported by The Centre for Large AI Models (CLAIM) of The Hong Kong Polytechnic University.}
\thanks{Yiyang Jiang, Jiaxin Wu, Qing Li, and Xiao-Yong Wei are with the PolySmart Group, Department of Computing, The Hong Kong Polytechnic University, Hong Kong, China (e-mail: yiyang.jiang@connect.polyu.hk; nikki-jiaxin.wu@polyu.edu.hk; qing-prof.li@polyu.edu.hk; cs007.wei@polyu.edu.hk).}
\thanks{Yiyang Jiang, Guangwu Qian, Qi Huang, and Xiao-Yong Wei are with the College of Computer Science, Sichuan University, Sichuan, China (e-mail: cswei@scu.edu.cn).}
\thanks{Yongkang Wu is with the Department of Laboratory Medicine and
Outpatient, West China Hospital, Sichuan University, Sichuan, China (e-mail: vipwyk@163.com).}
\thanks{Yiyang Jiang and Guangwu Qian contributed equally to this work}
\thanks{Xiao-Yong Wei is the corresponding author.}
}

\maketitle

\begin{figure*}[htbp]
\centering
\subfigure[Different fluorescence patterns of cell-level images]{
\includegraphics[width=1\textwidth]{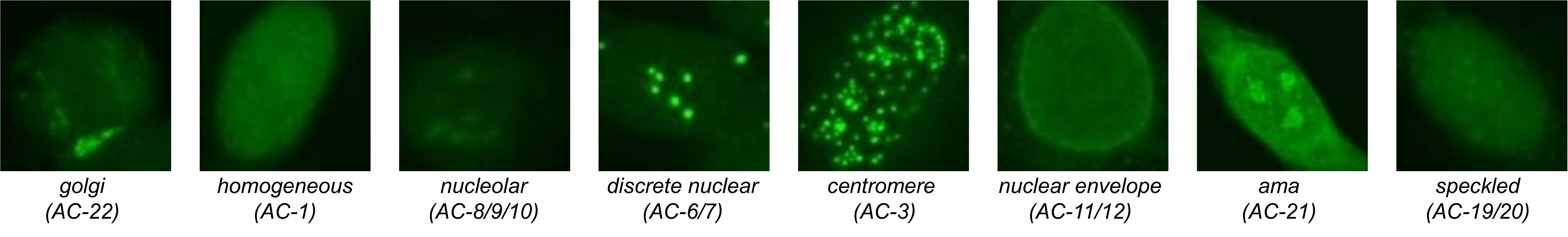}
\label{fig:cell-level}
}
\subfigure[Different fluorescence patterns of specimen-level images]{
\includegraphics[width=1\textwidth]{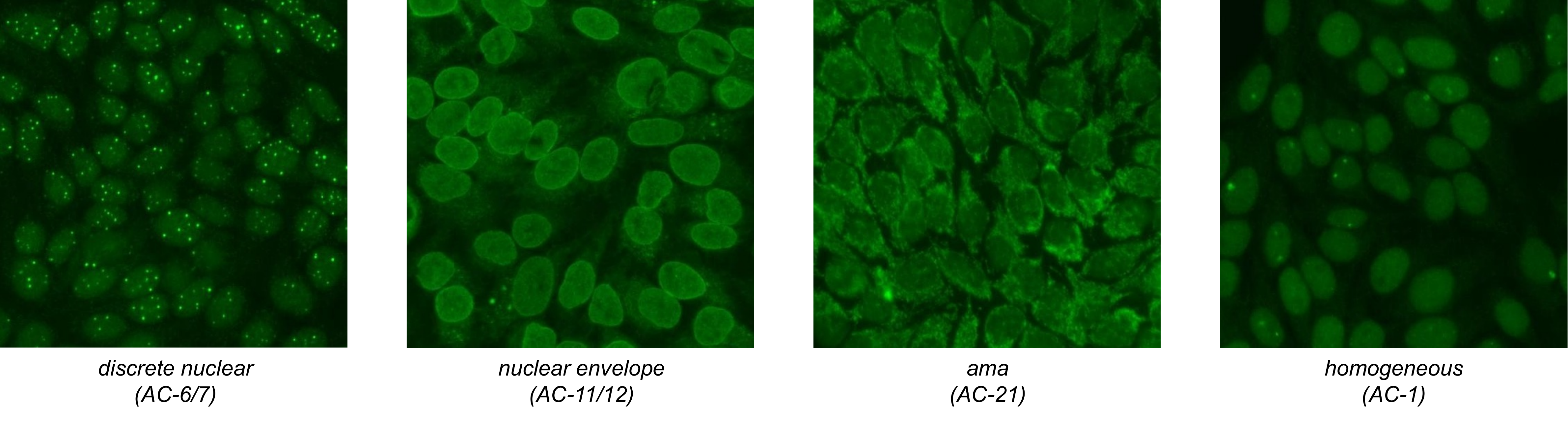}
\label{fig:specimen-level}
}
\subfigure[ANA images of multiple fluorescence patterns under clinical setting]{
\includegraphics[width=1\textwidth]{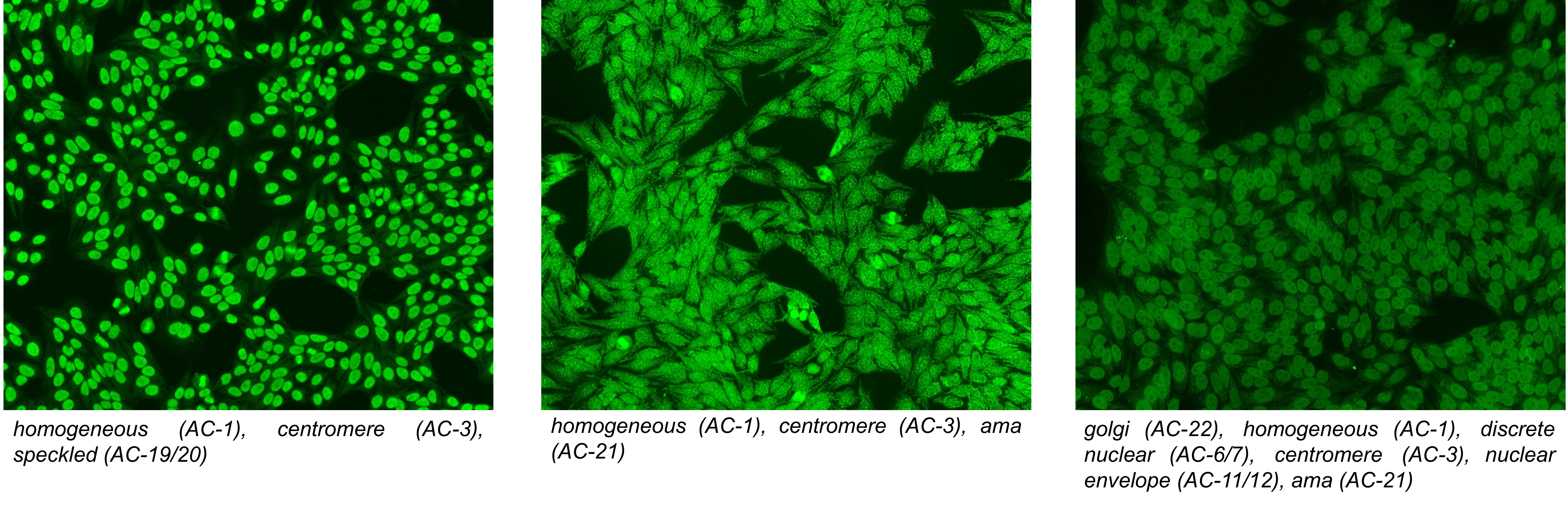}
\label{fig:clinical}
}
\caption{Examples of ANA images and visualization of the multi-label multi-instance challenge.}
\end{figure*}

\begin{abstract}
Antinuclear antibody (ANA) testing is a critical
method for diagnosing autoimmune disorders such as
Lupus, Sjögren’s syndrome, and scleroderma. 
Despite its importance, manual ANA detection is slow, labor-intensive, and demands years of training. ANA detection is complicated by over 100 coexisting antibody types, resulting in vast fluorescent pattern combinations. 
Although machine learning and deep learning have enabled automation, ANA detection in real-world clinical settings presents unique challenges as it involves multi-instance, multi-label (MIML) learning. In this paper, a novel framework for ANA detection is proposed that handles the complexities of MIML tasks using unaltered microscope images without manual preprocessing. Inspired by human labeling logic, it identifies consistent ANA sub-regions and assigns aggregated labels accordingly. These steps are implemented using three task-specific components: an instance sampler, a probabilistic pseudo-label dispatcher, and self-paced weight learning rate coefficients. 
The instance sampler suppresses low-confidence instances by modeling pattern confidence, while the dispatcher adaptively assigns labels based on instance distinguishability. Self-paced learning adjusts training according to empirical label observations. Our framework overcomes limitations of traditional MIML methods and supports end-to-end optimization. 
Extensive experiments on one ANA dataset and three public medical MIML benchmarks demonstrate the superiority of our framework. On the ANA dataset, our model achieves up to +7.0\% F1-Macro and +12.6\% mAP gains over the best prior method, setting new state-of-the-art results. It also ranks top-2 across all key metrics on public datasets, reducing Hamming loss and one-error by up to 18.2\% and 26.9\%, respectively.
The source code can be accessed at \href{https://github.com/fletcherjiang/ANA-SelfPacedLearning}{https://github.com/fletcherjiang/ANA-SelfPacedLearning}.
\end{abstract}


\begin{figure*}
    \centering
    \includegraphics[width=1\linewidth]{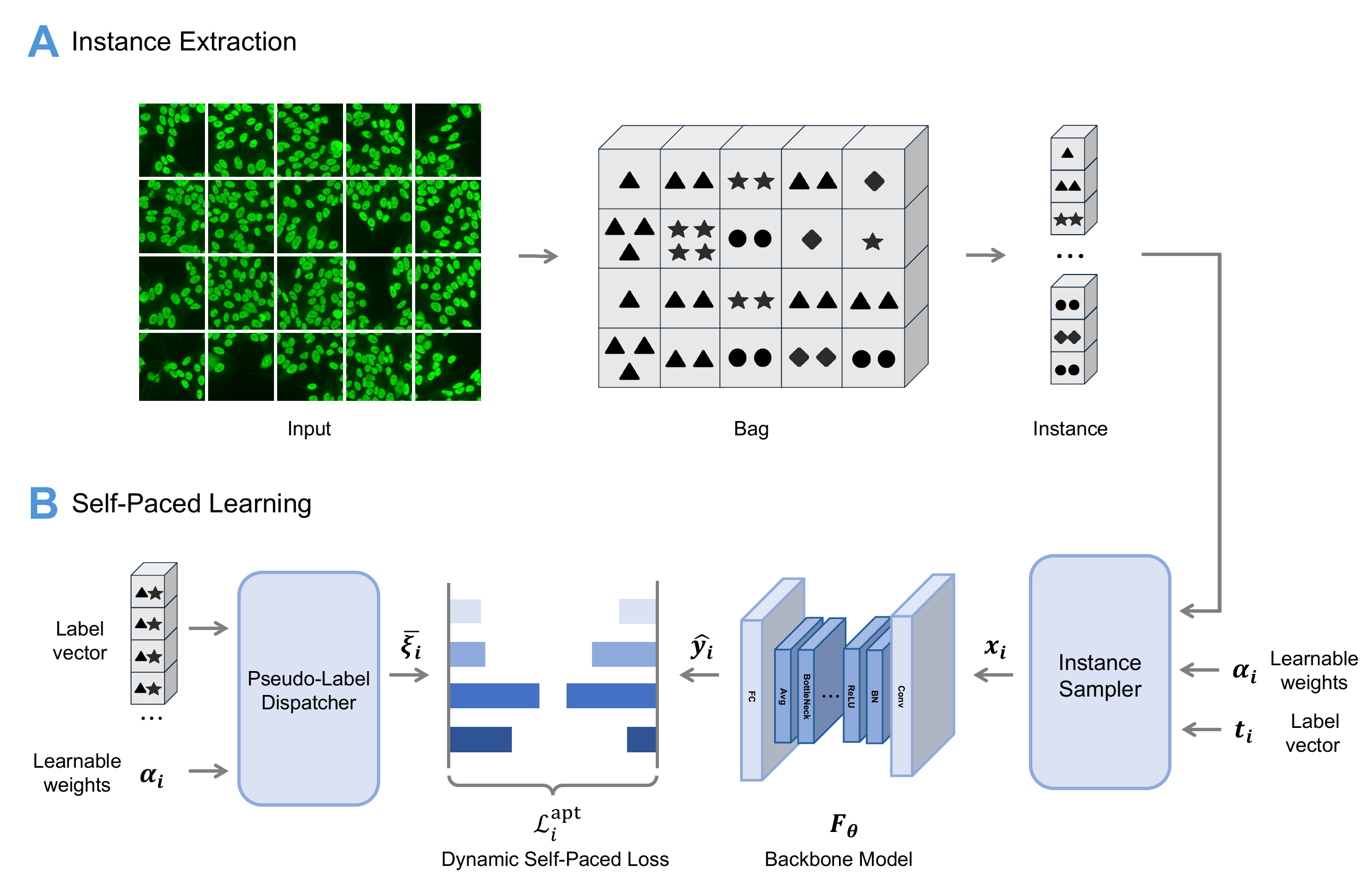}
    \caption{Overview of the proposed architecture, the framework begins by extracting instances from an input ANA image, different symbols (\ding{115}, \ding{108}, \ding{79}, \ding{117}) represent different ANA patterns detected in sub-regions of the ANA image. Each cell in the grid represents a local patch containing one or more ANA patterns. These instances are then evaluated by an Instance Sampler, which leverages learnable confidence weights to select the most representative instances for training. A CNN model provides initial predictions on the chosen instances, and a Pseudo-Label Dispatcher generates soft, continuous pseudo-labels based on both the image-level labels and the learnable confidence weights. The Dynamic Self-Paced Loss uses these pseudo-labels to adaptively emphasize more reliable instances and guide the network’s learning process. Finally, all subregion predictions are aggregated for a new image to produce the final multi-label ANA classification result.}
    \label{fig:anamodel}
\end{figure*}

\vspace{4em}
\section{Introduction}
\label{sec:introduction}
\IEEEPARstart{A}{ntinuclear} antibody (ANA) test is a widely used diagnostic method for a variety of autoimmune disorders, including systemic lupus erythematosus, Sj\"ogren syndrome, scleroderma, mixed connective tissue disease, polymyositis, dermatomyositis, autoimmune hepatitis, and drug-induced lupus. 
These disorders can progress to life-threatening conditions, such as cancer, or lead to death, making ANA detection a critical and longstanding focus in the medical field. 
In this work, a \emph{pattern} denotes the characteristic morphology and subcellular distribution of fluorescence on HEp-2 cells, specifying which compartments (e.g., nucleus, nucleolus, centromeres, cytoplasm) exhibit signal and in what configuration (e.g., homogeneous, speckled, nucleolar), as determined by autoantibody binding to specific antigens. It does not refer to periodic texture in the sense of computer vision.

In the ANA test, patient serum is incubated on HEp-2 cells, and a fluorescent secondary antibody is applied to reveal nuclear binding. The resulting staining pattern and titer indicate ANA positivity, guiding interpretation.
Recently, ANA detection has also drawn significant attention from the medical image recognition community, spurred in part by efforts from ICPR (International Conference on Pattern Recognition) since 2012.
ICPR introduced a series of ANA datasets between 2012 and 2016 as benchmarks for ANA detection challenges. 
These datasets typically consist of manually selected and labeled images, each containing either a single cell or multiple cells with the same pattern (as shown in Fig.~\ref{fig:cell-level} and Fig.~\ref{fig:specimen-level}).
Consequently, most prior studies have formulated ANA detection as a multi-class classification problem for either single-instance or multi-instance images \cite{6460100, 6460660, electronics8010020, LEI2018290, electronics8080850, 7899611, 8545751, 7899614, 7950689, 7899613, 7400923, 7899953, MANIVANNAN201612,xiao2024confusion,jiang2024prior,xiao2025curiosity,xiao2025diffusion}.
However, this approach is impractical in real-world clinical or laboratory settings. 
Technicians typically sample representative sub-regions from microscope images for ANA labeling. 
These sub-regions often contain hundreds of cells, potentially with different ANA patterns co-occurring (e.g., AC-1 and AC-3 in Fig.~\ref{fig:clinical}). 
The reasoning behind using sub-regions for detection is that antibodies are produced in response to antigens, whose heterogeneous distribution across cells leads to uneven antibody binding; analyzing sub-regions captures this variability.
To ensure reliable detection, technicians examine sub-regions to confirm whether antibodies are consistent across most cells, rather than occurring by chance in isolated instances.

Detecting antinuclear antibodies (ANAs) is a challenging task, even for experienced technicians.
The difficulty arises from the vast number of antibody types and their tendency to coexist within cells rather than appearing in isolation. 
This coexistence means that the fluorescent pattern observed is determined by the combination of ANA types.
%
Given roughly $100$ ANA pattern types, the space of multi-label assignments is $2^{100} (\approx 10^{30})$, because each pattern is independently present or absent.
This combinatorial effect further complicates pattern detection.
Several international efforts have been made to standardize and define patterns, such as the recommendations by the European Autoimmunity Standardization Initiative \cite{Agmon-Levin17} and the International Consensus on ANA Patterns (ICAP) \cite{4abf7eaac7734f98a32b7591364aab99}.
Despite these efforts, only 29 patterns (labeled AC-1 to AC-29) have been officially recognized and agreed upon by ICAP, which is a small subset of the theoretical pattern space. 
This discrepancy makes detection prone to distractions caused by the many unrecognized patterns.
As a result, it takes years of rigorous training for a technician to become proficient in detecting and interpreting ANA patterns accurately.

From a machine learning perspective, ANA detection can be framed as a multi-instance multi-label (MIML) problem \cite{NIPS2006_8e489b49}, as ANA samples consist of multiple sub-regions within an image, each associated with mixed labels. 
This makes it one of the most challenging learning problems.
In prior research, multi-instance learning \cite{NIPS1990_e46de7e1, DIETTERICH199731} and multi-label learning \cite{6471714} have typically been studied independently. 
However, the challenges in MIML stem not only from the combination of these two learning paradigms but also from the incompatibility of ANA data with the common assumptions underlying multi-instance or multi-label learning methods.
First, as illustrated in Fig.~\ref{fig:learning_frameworks}(b), most multi-instance learning approaches are designed for instances that share a consistent label, often framed as a binary classification task (e.g., distinguishing foreground from background) \cite{8822590, wei2021deep}. 
The presence of mixed instance (sub-region) labels in ANA images renders these traditional multi-instance models unsuitable.
Additionally, multi-label learning is typically applied to image datasets (e.g., Core15K/16K \cite{10.1007/3-540-47979-1_7}, NUS-WIDE \cite{10.1145/1646396.1646452}, Scene \cite{1467486}, IMDB \cite{10.1007/978-3-642-04174-7_17}, MS-COCO \cite{10.1007/978-3-319-10602-1_48}, WSS4LUAD \cite{Han2022WSSS4LUADGC}) where each label is often associated with a few distinct instances. 
Importantly, these instances typically represent natural objects (e.g., cats, cars) with clear, distinctive appearances. 
In contrast, ANA sub-regions exhibit arbitrary shapes and highly non-distinctive appearances, making classification far more complex.
The difficulty of this task, as highlighted by the international standardization efforts led by ICAP, further underscores the challenges inherent to ANA detection, even for human experts.

In this paper, we present a pilot study that explores the use of multi-instance multi-label (MIML) learning for ANA recognition in a real-world clinical and laboratory setting. 
Unlike previous studies, the samples we use are unaltered microscope images without any manual preprocessing, such as cropping or resampling.
As shown in Fig.~\ref{fig:anamodel}, we propose a prototype framework inspired by the two-step logic employed by human technicians when labeling: 1)
Identify sub-regions with consistent ANA presence (and disregard the rest). 2) Assign labels to the identified sub-regions and aggregate these labels at the sample level.
The first step is implemented using an \textbf{instance sampler}, which dynamically models and updates the learner's confidence regarding the presence of ANAs across training epochs. 
This confidence is used to suppress the influence of less confident instances during both the input and back-propagation stages, leveraging \textbf{instance-level attention} and a \textbf{dynamic self-paced loss} collaboratively.
The second step is implemented through a \textbf{probabilistic pseudo-label dispatcher} and \textbf{label-aware learning rate coefficients}. 
The dispatcher integrates instance-level distinguishability, adaptively distributing the aggregated sample-level labels to individual instances. 
Meanwhile, the label-aware coefficients regulate the learning process by incorporating empirical observations about label distributions and ensuring a controlled and adaptive training procedure.
The framework’s key advantage lies in addressing MIML challenges in ANA detection through the use of self-paced learning. 
All parameters are automatically optimized in an end-to-end fashion.\\

\section{Related Work}
\subsection{ANA Detection}
\label{sec:ANA}
Autoimmune disorders are diseases that happen when the immune system loses its ability to tolerate the body’s own tissues and causes damage to organs and tissues through autoantibodies or sensitized lymphocytes. 
Antinuclear antibodies (ANAs) are among the most common autoantibodies in these disorders, making their detection necessary for diagnosis. 
In manual detection, doctors examine ANA images under a fluorescence microscope and identify their patterns. 
This method is slow, requires much effort, and the results often depend on the doctor’s experience.
To address these issues, researchers have developed automatic ANA detection methods, which have gained significant attention.
Based on datasets, previous studies can be categorized into cell-level~\cite{6460100,6460660,7950689} and specimen-level ANA detection~\cite{li2016hep,manivannan2016automated}. 

Most methods focus on cell-level detection, which involves analyzing single-cell images.
In the early stages of machine learning, researchers extracted handcrafted features, combined them, and used machine learning classifiers~\cite{6460100,6460660}.
For example, Ghosh et al. \cite{SNELL20142338} converted raw images to grayscale, extracted features like shape, texture, and gradient histograms, and used Support Vector Machines (SVM) \cite{10.1023/A:1022627411411} for classification. 
Thibault et al. extracted morphological and statistical texture features and classified them using logistic regression, random forest, and neural networks \cite{6460660}.
With the rise of deep learning, many Convolutional Neural Network (CNN)-based methods have been proposed for single-cell ANA detection~\cite{7899611,7950689}. 
Gao developed a 3-layer CNN, which outperformed methods relying on handcrafted features \cite{7400923}. Building on ResNet \cite{7780459}, DSRN further improved performance by incorporating both feature maps and final outputs into the model \cite{8545751}.

Specimen-level ANA detection requires extra annotations to create a segmentation mask for each cell, which converts the task from specimen-level to cell-level. 
After predicting results for individual cells, an aggregation algorithm is used to combine these predictions into a specimen-level result. 
For example, Foggia et al. proposed a method that identifies the most common ANA type among cells and uses it as the specimen-level prediction \cite{ALDULAIMI2019534}. 
Li et al. proposed a different approach in
which all cell-level predictions were used to create a pattern histogram, which was then fed into an SVM to predict the specimen-level label \cite{7899953}. 
However, these methods, which depend on extra annotations, are not suitable for clinical practice where only sample-level labels are available.

\subsection{Multi-Instance Multi-Label Learning}
\label{sec:MIML}
\begin{figure}[htbp]
\centering
\includegraphics[width=0.48\textwidth]{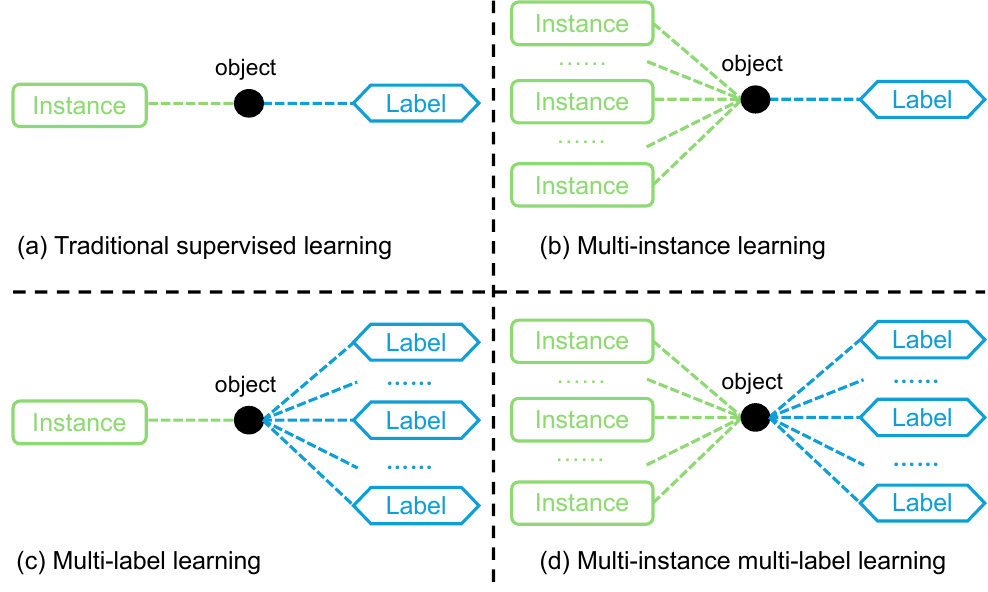}
\caption{Four different learning frameworks}
\label{fig:learning_frameworks}
\end{figure}

Most multi-instance learning methods are designed for binary or multi-class classification. 
However, ANA detection in clinical practice is a typical multi-instance multi-label (MIML) task, which requires specialized algorithms. 
The differences between supervised learning, multi-instance learning, multi-label learning, and MIML are shown in Fig.~\ref{fig:learning_frameworks}.
Among these, MIML is the most challenging because it combines the complexities of both multi-instance learning and multi-label learning.

Zhou et al. proposed two approaches to solve MIML tasks by converting them into either multi-instance learning or multi-label learning tasks \cite{NIPS2006_8e489b49}. 
Based on these approaches, he developed MIMLBoost and MIMLSVM to handle the tasks effectively. 
Inspired by advancements in deep learning, Feng introduced the DeepMIML network, which eliminates the need for pre-generating instances \cite{10.5555/3298483.3298513}. 
Instead, it automatically creates instance representations using a deep neural network structure. 
DeepMIML also includes a sub-concept layer designed to uncover hidden connections between input patterns and output semantic labels.

\subsection{Self-Paced Learning}
\label{sec:self-paced learning}
In 2010, Kumar et al. introduced Self-Paced Learning (SPL) \cite{NIPS2010_e57c6b95} as an extension of curriculum learning (CL), which Bengio proposed in 2009 \cite{10.1145/1553374.1553380}. 
SPL is based on the idea of mimicking how humans learn: starting with simple examples and gradually moving to more complex ones \cite{NIPS2011_f9028fae}. 
Specifically, SPL addresses dataset issues such as class imbalance and corrupted labels by adding a pace-controlled regularization term to the learning objective. A minimal form is:
\begin{equation}
\label{eq:spl-brief}
\min_{\theta,\;\mathbf{w}\in[0,1]^n}
\sum_{i=1}^{n} \Big[ w_i\mathcal{L}(y_i,\hat{y}_i) + r(w_i, c) \Big],
\end{equation}
where $\hat{y}_i$ denotes the model prediction for $x_i$, $\mathcal{L}$ is any supervised loss, and $r(\cdot; c)$ is the self-paced regularizer with pace $c>0$. Here, $w_i\in[0,1]$ is the self-paced weight assigned to sample $i$ and is optimized jointly with $\theta$.
Intuitively, $w_i$ increases for “easy” samples (low loss) and decreases for hard/noisy ones, with the pace $c$ gradually relaxing the selection.
Our proposed framework is inspired by self-paced learning, but we have developed task-specific components to enhance the performance.
There are several variations of the SPL learning framework, but we can only briefly cover those designed for multi-instance learning or multi-label learning, as our focus is on MIML. 
Zhang et al. applied self-paced learning to multi-instance learning and replaced binary weights with real-valued ones, which were directly calculated using the loss function \cite{7410432, 7469327}. 
Li et al. extended self-paced learning to multi-label learning by considering the relationships between labels \cite{7929370}. 
Instead of assigning weights based on loss values, Zhong introduced a new reweighting function that considers instance complexity, determined by the distance between instance features and their corresponding labels \cite{ZHONG2021428}. 
Ren et al. proposed meta-learning algorithms guided by a small amount of unbiased validation data, assigning weights to training examples based on gradient directions \cite{pmlr-v80-ren18a, 10.5555/3454287.3454459}.
Our framework, in contrast, does not explicitly calculate weights or rely on unbiased validation data. 
Instead, it leverages prior knowledge about multi-label reliability and introduces label-level weights, which are used by three enhanced components to build an end-to-end deep model.

\section{Methodology}
\label{sec:methodlogy}
\subsection{Problem Formulation}
\label{sec:problem}
We formulate the ANA task as a multi-instance multi-label (MIML) learning problem, and an overview of the proposed framework is shown in Fig.~\ref{fig:anamodel}. 
Given a set of unaltered microscope ANA images and their ground truth labels $\{(x_i,y_i)\}^n_{i=1}$, the goal is to learn a function that takes an image $x_i$ as input and predicts $\hat{y_i}$ to best match $y_i$. Let ${L}=\{l_1,l_2,\dots,l_K\}$ be the set of predefined ANA pattern labels (e.g., Golgi, Nucleolar), where each $l_k$ denotes a specific pattern.
%

To address the complexity of image-level prediction in ANA classification, we adopt a divide-and-aggregate strategy. 
Specifically, each ANA image is decomposed into a bag of instances, denoted as $x_i = \{ x_i^1, x_i^2,\dots,x_i^{N_i}\}$, where \( N_i \) is the number of extracted instances from image \( x_i \). 
We denote by $i$ the \textbf{image/bag} index, by $j$ the \textbf{instance/patch} index within image $x_i$, and by $k$ the \textbf{class/label} index.
Let $x_i$ be the input image and $x_i^{j}$ its $j$-th instance. 
Instance-level predictions are first made independently, and then aggregated to produce the final image-level prediction. 

Instead of uniform sampling, we introduce \textit{\textbf{an instance sampler}} to extract meaningful sub-regions from ANA images.
Let ${t}_i^j$ denote a 0--1 label vector, where the $k^{th}$ element $t_{i,k}^j$ is equal to $1$ if $l_k$ is present in $x_i^j$, and $0$ otherwise.
The general idea is to find a vector $\boldsymbol \alpha_i^j=\{\alpha_{i,k}^j\}$ for each instance ${x}_i^{j}$, where the $k^{th}$ element $\alpha_{i,k}^j$ indicates how confident the learner is about the presence of $l_k$ in the instance.
This sampler is designed to assign a confidence score to each candidate region and filter out irrelevant or uninformative areas (e.g., regions without ANA cells). This step ensures that only high-quality instances are used for downstream prediction.

The learning process can thus be formulated as a two-stage inference:
\begin{equation}
\hat{y}_i^j = \mathcal{F}\big(\mathcal{S}(x_i^j,\alpha_{i,k}^{j},t_{i,k}^{j})\big),
\end{equation}
\begin{equation}
\hat{y}_i=\mathcal{G}\left(\{\hat{y}_i^{j}\}_{j=1}^{N_i}\right).
\end{equation}
where $\mathcal{S}(\cdot)$ is the instance sampler function,  $\mathcal{F}(\cdot)$ is the instance-level prediction function, and $\mathcal{G}(\cdot)$ denotes the aggregation function that synthesizes instance-level outputs into an image-level prediction $\hat{y}_i$.
The learning objective minimizes the instance-level binary cross-entropy between $\hat y_i^{j}$ and the pseudo-label $\bar{\xi}_i^{j}$ produced by $\mathcal{D}$ (Sec.~\ref{sec:pseudo-label}).

However, only image-level ground-truth labels $y_i$ are available in real-world scenarios, and instance-level annotations $y_i^j$ are typically not provided.
To address this challenge, we introduce \textit{\textbf{a probabilistic pseudo-label dispatcher}}, which estimates a soft pseudo-label \emph{vector} $\bar{\xi}_i^{j} =\{\bar{\xi}_{i,k}^{j}\}$ for each instance based on the image-level annotation:
\begin{equation}
\label{eq:dispatcher}
\bar{\xi}_{i,k}^{j} = \mathcal{D}\big(\alpha_{i,k}^{j}, t_{i,k}^{j}\big).
\end{equation}
Here $y_i\in\{0,1\}^K$ is the image-level multi-hot ground-truth label. We set $t_i \equiv y_i$ and denote its broadcast to instance $j$ by $t_i^{j}:=t_i$ for all $j\in\{1,\dots,N_i\}$. 
In other words, $t_i^{j}$ is the per-instance copy of $y_i$ and the $\alpha_{i,k}^j$ denotes the learnable per-instance, per-class weight vector. 
We associate each instance $x_i^{j}$ with a per-class confidence vector
$\boldsymbol \alpha_i^j$.
Both the instance sampler and the probabilistic pseudo-label
dispatcher are built upon $\boldsymbol \alpha_i^j$.
The concrete computation of $\mathcal{D}$ is detailed in Eq.~\ref{eq_pseudo-label}.
The instance-level objective minimizes the binary cross-entropy between the predicted probabilities and the pseudo-labels. Furthermore, we also introduce \textit{\textbf{a self-paced loss function}} $\mathcal{L}(\cdot)$ to dynamically adjust the contribution of each instance's loss to the overall objective.

In the following subsections, we illustrate our implementation of the instance-level prediction function $\mathcal{F}(\cdot)$, the aggregation function $\mathcal{G}(\cdot)$, the instance sampler $\mathcal{S}(\cdot)$, the probabilistic pseudo-label dispatcher  $\mathcal{D}(\cdot)$, and the self-paced loss function~$\mathcal{L}(\cdot)$.

\subsection{Instance Sampler}
\label{sec:sampler}
We design the instance sampler $\mathcal{S}(\cdot)$ to select informative sub-regions from each ANA image, imitating the technician's strategy of identifying the most representative regions for ANA pattern recognition. 
To construct instances for multi-instance learning, we adopt a grid-based sliding window strategy to partition each input image $x_i$ into $N_i$ non-overlapping sub-regions, denoted as ${x}_i^j$. Each sub-region is a fixed-size patch of $448 \times 448$ pixels.

The idea is to encourage the selection of sub-regions in which the learner has developed higher confidence (from previous epochs) about ANA presence, while suppressing less confident ones.

\begin{equation}
\label{eq_sampling}
s_i^{j}=\max_{k}\Big[\mathrm{ReLU}\big(\alpha_{i,k}^{j}\big)t_{i,k}^{j}\Big],\qquad \text{with } t_i^{j}\equiv t_i.
\end{equation}
where $\mathrm{ReLU}(\cdot)$ avoids negative weights. The per-instance sampling score $s_i^{j}$ is the maximum (over classes) of the masked weights.

Then, the training instance ${x}_i^j$ is sampled from the image using a weighted random sampling strategy based on $s_i^j$, which reflects the model's confidence in that region. In other words, instances ${x}_i^j$ with higher $s_i^j$ values (indicating higher confidence of containing relevant labels) are more likely to be selected and contribute to training.
Instead of employing a summation or mean, the maximum is preferred due to two reasons: 1) Even if only a minority of ANA types are given high confidences, i.e., a distinguishable instance with dominating ANA types, the maximum strategy still ensures such an instance can be easily sampled.
2) When all ANA types of an instance share similar confidences, i.e., a hard instance with non-dominating ANA types, although the maximum strategy sounds like a random one, the sampling probability is also low.
Only confidences on labeled ANA types are considered, because we believe that ground truths labeled by technicians at the sample level are always correct, but distributing labels directly to sub-region instances is inappropriate. Instead of using the full set of sample-level labels, a subset obtained by setting certain elements of  $t_i^j$ from 1 to 0 is a more appropriate candidate for instance-level labels.

\subsection{Instance-Level Prediction and Aggregation}
After sampling the informative sub-regions $x_i^{j}$ from image $x_i$, each instance $x_i^{j}$ is passed through a shared prediction network to produce instance-level classification scores:

\begin{equation}
\hat{{y}}_i^j = \mathcal{F}({x}_i^j, \theta),
\end{equation}
where $\hat{{y}}_i^j \in [0,1]^K$ denotes the predicted class probabilities for the $j$-th instance, and $\theta$ represents the parameters of a CNN-based backbone (e.g., ResNet).

To produce the final image-level prediction, we aggregate the instance-level outputs using an aggregation function $\mathcal{G}(\cdot)$ that operates in two steps: class-wise max pooling followed by thresholding. Specifically, for each class $k$, the aggregated prediction is given by:
\begin{equation}
\label{eq_aggregation}
\hat{y}_{i,k} = \mathbbm{1} \left( \max_j \hat{y}_{i,k}^{j} > \tau \right),
\end{equation}
where $\hat{y}_{i,k}$ is the predicted probability of class $k$ for instance $j$, $\tau$ is a fixed threshold (e.g., 0.5), and $\mathbbm{1}(\cdot)$ is the indicator function. This formulation reflects the multi-instance learning assumption: if any sub-region strongly supports a specific ANA pattern, the entire image is considered positive for that label.

\subsection{Probabilistic Pseudo-label Dispatcher}
\label{sec:pseudo-label}
In the absence of instance-level annotations, we introduce a probabilistic pseudo-label dispatcher $\mathcal{D}(\cdot)$ to generate soft supervisory signals for each instance.
The probabilistic pseudo-label dispatcher distributes pseudo-labels based on the accumulated (sample-level) labels and the corresponding probabilities. 

The weighting factor $\boldsymbol \alpha_i^j=\{\alpha_{i,k}^j\}$ is leveraged to generate pseudo-labels, eliminating the need to introduce additional variables.
For each instance, a scalar weighting factor is sufficient for the instance sampler; however, such granular weighting factors are incapable of pseudo-label adjustment. This motivates the design of a vector $\boldsymbol \alpha_i^j$ for each instance $x_i^j$ from the outset. 
Furthermore, pseudo-labels for each instance, such as predictions $\hat{y}_{i}^j$, are non-negative real numbers in the range $[0,1]$, while ground truths are 0--1 label vectors $t_i^j$. Given the learnable weighting factor $\boldsymbol \alpha_i^j$, the pseudo-label $\bar{\xi}_{i,k}^{j}$ is defined as follows:




\begin{equation}
\label{eq_pseudo-label}
\mathcal{D}\big(\alpha_{i,k}^{j}, t_{i,k}^{j}\big) =
\left\{
\begin{aligned}
\xi_{i,k}^{j} &= \mathrm{ReLU}(\alpha_{i,k}^{j}) t_{i,k}^{j} \\[2pt]
\bar{\xi}_{i,k}^{j} &= \dfrac{\xi_{i,k}^{j} - \min\limits_{k}\xi_{i,k}^{j} }
{\max\limits_{k} \xi_{i,k}^{j} - \min\limits_{k}  \xi_{i,k}^{j}}
\end{aligned}
\right. ,
\end{equation}
where $\bar{\xi}_{i,k}^{j}$ is the max--min normalized $\xi_{i,k}^{j}$. 
Element-wise multiplication by ground truths constrains generated pseudo-labels in subsets of accumulated labels, while max--min normalization ensures pseudo-labels in the range $[0,1]$. 
Rather than 0--1 vectors, pseudo-labels in the range $[0,1]$ are directly regarded as targets in the loss function, because real numbers can precisely reflect the level of confidence of the generated pseudo-labels. 
Instead of ground truths, pseudo-labels are generated and used for supervision in the training stage, while only ground truths are employed for evaluation and testing procedures.

\subsection{Self-Paced Loss Function}
In case CNN is selected to implement the prediction function, the learning is conducted by adaptive moment estimation (Adam) for minimizing the cross-entropy loss. The baseline image-level binary cross-entropy is:

\begin{equation}
\label{eq:img_ce}
\mathcal{L}_{i}
= -\sum_{k=1}^{K}\Big[t_{i,k} \log \hat y_{i,k} + \big(1-t_{i,k}\big)\log\big(1-\hat y_{i,k}\big)\Big],
\end{equation}
\\
In real-world settings, only image-level labels are available (no instance-level annotations). 
We therefore introduce an instance sampler to select informative patches and a pseudo-label generator to enable instance-level training. Both the instance sampler and the pseudo-label generator are built upon $\boldsymbol \alpha_i^j$. Finally, $\boldsymbol \alpha_i^j$ is used to control the contribution of the instance in learning, which results in a final dynamic self-paced loss function:
\begin{equation}
\label{eq:inst_loss}
\begin{aligned}
\mathcal{L}_i^{\text{apt}}
&= -\sum_{k=1}^{K}\Big[
\alpha_{i,k}^{j}\bar{\xi}_{i,k}^{j} \log(\hat{y}_{i,k}^{j}) \\
&\qquad
+ \big(1-\bar{\xi}_{i,k}^{j}\big)\log\big(1-\hat{y}_{i,k}^{j}\big)
\Big] \\
&= -\Big[
(\alpha_i^{j}\odot\bar{\xi}_i^{j})^{\top}\log \hat{y}_i^{j}
+ (\mathbf{1}-\bar{\xi}_i^{j})^{\top}\log(\mathbf{1}-\hat{y}_i^{j})
\Big],
\end{aligned}
\end{equation}
where $\odot$ is the Hadamard product.
The key challenge is to design strategies for constructing and updating the two vectors $\boldsymbol \alpha_i^j$ and $\bar{\xi}_i^j$, which are the results of the instance sampler and the pseudo-label generator, respectively.

\section{Experiments}
\label{sec:exp}

\subsection{Datasets}
\label{sec:dataset}
We evaluate our proposed method from two perspectives: its effectiveness in ANA detection and its generalizability to other multi-label medical tasks.
The ANA detection experiments are conducted on a real-world ANA dataset collected through the Integrated Care Organization (WCO), which comprises 686 hospitals across West China.
The data aggregation process is led by West China Hospital, Sichuan University.
The dataset contains 6,563 ANA immunofluorescence images, each labeled by clinicians during the diagnostic process, covering 8 ANA pattern classes. 
Labels were subsequently cleaned, corrected, and verified by a panel of three expert technicians. 
A label was finalized as ground truth only if at least two technicians agreed.

The dataset is categorized into 3,359 single-label samples, 2,685 double-label samples, and 519 samples with more than three labels. 
From a category perspective, the samples are distributed as follows: 116 Golgi (AC-22), 1,770 homogeneous (AC-1), 682 nucleolar (AC-8/9/10), 279 discrete nuclear (AC-6/7), 613 centromere (AC-3), 373 nuclear envelope (AC-11/12), 2,546 AMA (AC-21), and 4,115 speckled (AC-19/20). 
Labels from different categories may coexist in the same sample.
To divide the dataset, 70\% of the samples were assigned to the training set, while the validation and test sets each received 15\%. 
For multi-instance experiments, visual patches were extracted from the samples using a sliding window algorithm, with each window sized $448 \times 448$. 
This process resulted in 82,240 patches from 4,605 training samples, 19,400 patches from 939 validation samples, and 19,330 patches from 939 test samples.

To validate the generalizability of our approach, we conduct experiments on three real-world multi-label medical image datasets involving \textit{whole-slide images} (WSIs):
\begin{itemize}
    \item \textbf{NuCLS}~\cite{amgad2021nucls}: 1,358 WSIs from TCGA breast cancer cohort, annotated with seven possible labels.
    \item \textbf{BCSS (Breast Cancer Semantic Segmentation)}~\cite{amgad2019structured}: 151 hematoxylin and eosin (H\&E) stained WSIs covering 22 histologically-confirmed breast cancer cases.
    \item \textbf{PaNNuke}~\cite{gamper2020pannuke}: 7,904 WSIs spanning 19 tissue types with five nuclear structure labels.
\end{itemize}
Each dataset is split into 60\% training, 10\% validation, and 30\% testing. Models are trained for 10 epochs with instance shuffling in each epoch.

\subsection{Implementation Details}
\label{sec:implementation}
All experiments are implemented in PyTorch \cite{NEURIPS2019_9015} and conducted on a computing platform equipped with an AMD EPYC 7542 CPU (2.9GHz), 94 GB of RAM, and an NVIDIA RTX 4090 GPU.
For the real-world ANA dataset, we adopt the Adam optimizer \cite{kingma20153rd} with $\beta=(0.9, 0.999)$, a learning rate of $5 \times 10^{-3}$, and a batch size of 32. All backbone networks are pre-trained on ImageNet \cite{5206848}, and input patches of size $448 \times 448$ are resized to match the input dimensions of the corresponding models.

For the three real-world multi-label medical image datasets, we follow a unified hyperparameter protocol across all competing methods unless otherwise specified in their original implementations. In our setup, each image is divided into smaller patches of approximately $64 \times 64$ pixels, with the total number of instances per image determined by its resolution. All models are trained and evaluated using 10-fold cross-validation to ensure robustness and fair comparison.

\subsection{Evaluation Metrics}
For the ANA task, we report four widely adopted metrics: $F1_{Micro}$, $F1_{Macro}$, accuracy, and mean Average Precision (mAP). Notably, $F1_{Micro}$ treats all instances equally, while $F1_{Macro}$ averages over all categories, allowing assessment from both global and per-class perspectives. 
An overall metric is also reported by averaging the above four scores.

For other medical tasks, we adopt four standard metrics commonly used in MIML literature: \textit{Hamming Loss} (HL), \textit{One Error} (OE), \textit{Ranking Loss} (RL), and \textit{Average Precision} (AP), following the definitions outlined in~\cite{10170751}.

\begin{table}[htbp]
\centering
\caption{Experimental results of SOTA methods and ours on the test set. The best results are bold. Models marked with $^{\dagger}$ are specifically designed for ANA detection.}
\label{tbl:sota}
\begin{tabular}{lccccc}
\toprule
\textbf{Model} & $F1_{\text{Micro}}$ & $F1_{\text{Macro}}$ & Acc. & mAP & Overall \\
\midrule
\rowcolor[gray]{.8} \multicolumn{6}{c}{Multi-label} \vspace{0.4em}
\\
Add-GCN~\cite{ye2020attentiondrivendynamicgraphconvolutional}  & 0.771          & 0.673          & 0.440          & 0.799             & 0.671 \\
ASL~\cite{benbaruch2021asymmetriclossmultilabelclassification}      & 0.667          & 0.568          & 0.159          & {\textbf{0.817}}    & 0.553 \\
MIMLmiSVM~\cite{zhou2012multi} & 0.483          & 0.088          & 0.243          & 0.145             & 0.240 \\ 
MIMLkNN~\cite{zhang2010k} & 0.455          & 0.079          & 0.256          & 0.156             & 0.237 \\ 
MIMLBoost~\cite{zhou2006multi} & 0.461          & 0.091          & 0.182          & 0.168             & 0.226\\ 
DeepMIML~\cite{10.5555/3298483.3298513} & 0.493          & 0.098          & 0.278          & 0.199             & 0.267 \\ 
MIMLfast~\cite{huang2018fast} & 0.513          & 0.355          & 0.478          & 0.677             & 0.506 \\ 
Q2L-CvT~\cite{liu2021query2labelsimpletransformerway} &0.582&  0.275&  0.324 & 0.433 & 0.404 \\
ML-Decoder~\cite{ridnik2021mldecoderscalableversatileclassification} &0.692&  0.373&  0.438 & 0.482 & 0.496  \\
CBAM~\cite{woo2018cbamconvolutionalblockattention}     & \underline{0.792}         & \underline{0.695}          & 0.471          & \underline{0.812}            & \underline{0.693} \\
IDA-SwinL~\cite{liu2023causality} &0.685&  0.380&  0.405 & 0.443 &  0.478\\
RRA$^{\dagger}$~\cite{zengRecognitionRareAntinuclear2024}  &0.747&  0.494&  \underline{0.505} & 0.603 & 0.587 \\
\textbf{Ours}     & {\textbf{0.821}} & {\textbf{0.745}} & {\textbf{0.573}} & 0.787             & {\textbf{0.732}} \\
\midrule
\rowcolor[gray]{.8} \multicolumn{6}{c}{Single-label} \vspace{0.4em} \\
FUS$^{\dagger}$~\cite{xie2023automatic} & 0.738          & 0.354          & \underline{0.738}          & 0.483             & 0.578 \\ 
RRA$^{\dagger}$~\cite{zengRecognitionRareAntinuclear2024}  &\underline{0.835}& \underline{0.711}& 0.698& \underline{0.872}& \underline{0.779} \\

\textbf{Ours}     & {\textbf{0.863}} & {\textbf{0.775}} & {\textbf{0.764}} & {\textbf{0.932}} & {\textbf{0.834}} \\

\bottomrule
\end{tabular}
\end{table}

\subsection{Comparison to SOTA Methods}
\label{sec:exp_comparison_SOTA}
To comprehensively evaluate the effectiveness of our approach, we compare it against a wide range of state-of-the-art (SOTA) methods on both the ANA dataset and three publicly available medical multi-label datasets, including NuCLS, BCSS, and PaNuke. The baselines include general-purpose multi-label classification models (e.g., CBAM~\cite{woo2018cbamconvolutionalblockattention}, Add-GCN~\cite{ye2020attentiondrivendynamicgraphconvolutional}, ASL~\cite{benbaruch2021asymmetriclossmultilabelclassification}, Q2L-CvT~\cite{liu2021query2labelsimpletransformerway}, ML-Decoder~\cite{ridnik2021mldecoderscalableversatileclassification}, IDA-SwinL~\cite{liu2023causality}), ANA-specific detection models (e.g., RRA~\cite{zengRecognitionRareAntinuclear2024}, FUS~\cite{xie2023automatic}), and representative MIML methods (e.g., MIMLNN~\cite{zhou2012multi}, MIMLSVM~\cite{zhou2006multi}, MIMLmiSVM~\cite{zhou2012multi}, MIMLkNN~\cite{zhang2010k}, MIMLBOOST~\cite{zhou2006multi}, MIMLfast~\cite{huang2018fast}, DeepMIML~\cite{10.5555/3298483.3298513}, BMIML~\cite{10170751}).\\

\subsubsection{Evaluation on ANA Dataset}
To ensure a fair and comprehensive evaluation, each method adopts its default backbone as specified in the original publications: ResNet-50 for CBAM, ASL, and our method; ResNet-101 for Add-GCN, IDA-SwinL, RRA, and FUS; VGG16 for DeepMIML; CvT-w24 for Q2L-CvT; and TResNet-L for ML-Decoder.

ANA images naturally show multi-label characteristics. Multiple fluorescence patterns can appear in the same cell. However, most previous ANA detection methods were designed and tested under single-label assumptions.
To enable a fair comparison, we additionally construct a single-label subset comprising 3,359 images annotated with only one ANA pattern. 
This allows benchmarking against single-label methods such as RRA and FUS.

As shown in Table~\ref{tbl:sota}, our method achieves state-of-the-art performance in both settings. 
In the multi-label scenario, our approach consistently outperforms all baselines in accuracy, $F1_{Micro}$, $F1_{Macro}$, and the overall averaged metric. 
Although ASL slightly surpasses our method in mAP, the difference is marginal. Meanwhile, our model's balanced performance across all metrics highlights its robustness and adaptability. DeepMIML ranks lowest, likely due to limitations in its shallow VGG16 backbone, lacking residual connections and multi-scale representation. 
Likewise, general-purpose models such as Q2L-CvT and ML-Decoder underperform, possibly because they were not tailored for the unique structural features of ANA data.

In the single-label setting, our model again surpasses RRA and FUS in all evaluation metrics, showcasing its strong generalization ability even under constrained labeling conditions. 
Given the prevalence of single-label classification in clinical ANA datasets, this result underscores the real-world applicability of our framework.\\

\begin{table*}[htbp]
\caption{Comparison results (mean $\pm$ std.) on three medical data sets\label{tab:table3}. $\uparrow$($\downarrow$) indicates that the larger (smaller) the value, the better the performance; Bold indicates the best performance of this metric; \underline{underline} indicates the next best performance of this metric; N/A represents that no result was obtained in 72 hours.}
\centering
\renewcommand{\arraystretch}{1.1}
\begin{tabular}{p{0.8cm}                      
  >{\centering\arraybackslash}p{1.8cm}    
  >{\centering\arraybackslash}p{1.5cm}
  >{\centering\arraybackslash}p{1.3cm}
  >{\centering\arraybackslash}p{1.3cm}
  >{\centering\arraybackslash}p{1.3cm}
  >{\centering\arraybackslash}p{1.3cm}
  >{\centering\arraybackslash}p{1.3cm}
  >{\centering\arraybackslash}p{1.3cm}
  >{\centering\arraybackslash}p{1.3cm}}
\bottomrule
Methods & MIMLNN          & MIMLSVM           & MIMLmiSVM       & MIMLkN   &MIMLBOOST          & MIMLfast  & DeepMIML & BMIML & \textbf{Ours}             \\ \hline
\rowcolor[gray]{.8} \multicolumn{10}{l}{\textit{NuCLS}}                                        \\ \hline
H.L.↓     & 0.125$\pm$0.004       & 0.106$\pm$0.008          & 0.494$\pm$0.017       & 0.233$\pm$0.005 & \underline {0.116$\pm$0.025}    & 0.253$\pm$0.028 &0.202$\pm$0.030          & 0.088$\pm$0.030 & \textbf{0.072$\pm$0.040}\\
O.E.↓     & 0.264$\pm$0.010       & 0.132$\pm$0.027          & 0.136$\pm$0.043       & 0.284$\pm$0.022 & \textbf{0.029$\pm$0.001} & 0.583$\pm$0.061 &0.525$\pm$0.008          & 0.037$\pm$0.015  & \underline{0.030$\pm$0.030}  \\
R.L.↓     & 0.077$\pm$0.002       & 0.041$\pm$0.020 & 0.368$\pm$0.017       & 0.380$\pm$0.023 & 0.099$\pm$0.005          & 0.392$\pm$0.004 &0.325$\pm$0.019          & \underline{0.043$\pm$0.010}   & \textbf{0.040$\pm$0.015} \\
A.P.↑     & 0.857$\pm$0.041       & 0.941$\pm$0.006    & 0.856$\pm$0.028       & 0.757$\pm$0.007 & 0.921$\pm$0.009          & 0.722$\pm$0.011 & 0.815$\pm$0.046         & \textbf{0.968$\pm$0.007}  & \underline{0.949$\pm$0.005} \\ \hline
\rowcolor[gray]{.8} \multicolumn{10}{l}{\textit{Breast}}                                                                                                           \\ \hline
H.L.↓     & 0.293$\pm$0.060 & 0.297$\pm$0.011          & 0.511$\pm$0.041       & 0.297$\pm$0.033 & 0.460$\pm$0.030          & 0.318$\pm$0.021 &0.541$\pm$0.032         & \underline{0.290$\pm$0.017} & \textbf{0.288$\pm$0.015} \\
O.E.↓     & 0.219$\pm$0.013       & 0.206$\pm$0.032          & 0.183$\pm$0.003       & 0.250$\pm$0.062 & \underline{0.013$\pm$0.001} & 0.500$\pm$0.016 &0.500$\pm$0.003         & 0.094$\pm$0.001   & \textbf{0.083$\pm$0.001} \\
R.L.↓     & {\underline 0.204$\pm$0.007} & 0.196$\pm$0.050          & 0.438$\pm$0.028       & 0.483$\pm$0.010 & 0.943$\pm$0.041          & 0.493$\pm$0.022 &0.502$\pm$0.046          & \underline{0.172$\pm$0.004} & \textbf{0.169$\pm$0.010}\\
A.P.↑     & 0.822$\pm$0.028       & .832$\pm$0.064    & 0.770$\pm$0.071       & 0.599$\pm$0.025 & 0.624$\pm$0.019          & 0.591$\pm$0.016 &0.530$\pm$0.026          & \textbf{0.854$\pm$0.021} & \underline{0.851$\pm$0.015} \\ \hline
\rowcolor[gray]{.8} \multicolumn{10}{l}{\textit{Pannuke}}                                                                                                          \\ \hline
H.L.↓     & 0.299$\pm$0.036       & 0.285$\pm$0.041    & 0.510$\pm$0.018       & 0.299$\pm$0.005 & N/A                & 0.377$\pm$0.011 & N/A      & \underline{0.276$\pm$0.005}  & \textbf{0.272$\pm$0.015} \\
O.E.↓     & 0.250$\pm$0.012       & \underline{0.167$\pm$0.024} & 0.182$\pm$0.033 & 0.200$\pm$0.022 & N/A                & 0.600$\pm$0.032 & N/A      & 0.212$\pm$0.038         & \textbf{0.155$\pm$0.017}  \\
R.L.↓     & 0.209$\pm$0.030       & \underline {0.189$\pm$0.006}    & 0.438$\pm$0.009       & 0.509$\pm$0.036 & N/A                & 0.465$\pm$0.031 & N/A      & \textbf{0.151$\pm$0.014}  & 0.197$\pm$0.016 \\
A.P.↑     & 0.806$\pm$0.042       & 0.823$\pm$0.045   & 0.770$\pm$0.013       & 0.441$\pm$0.040 & N/A                & 0.439$\pm$0.060 & N/A      & \textbf{0.846$\pm$0.003}  & \underline {0.836$\pm$0.005} \\ \bottomrule
\end{tabular}
\end{table*}

\subsubsection{Evaluation on  Public MIML Medical Datasets}
While many classical MIML methods, such as MIMLNN, MIMLSVM, MIMLmiSVM, MIMLkNN, MIMLBOOST, MIMLfast, and BMIML, rely on pre-extracted and fixed instance-level features, only DeepMIML adopts an end-to-end neural network backbone to learn instance representations jointly with bag-level prediction. Traditional approaches often depend on handcrafted features or simple image statistics, and subsequently apply shallow models like kernel-based classifiers, boosting frameworks, or k-nearest-neighbor algorithms. Though computationally lightweight and practical for small-scale data, these methods lack the capacity to capture rich visual semantics and are inherently limited when applied to high-resolution medical images or large datasets, where joint optimization is critical.

In contrast, our method fully leverages deep feature learning and achieves significantly stronger results across diverse and complex domains. As summarized in Table~\ref{tab:table3}, we evaluate our framework on three challenging public MIML medical datasets: NuCLS, BCSS, and PaNuke. Our method achieves the best or second-best performance in nearly all evaluation metrics: it attains the lowest Hamming Loss and Ranking Loss on NuCLS, the best One Error and Accuracy on BCSS, and maintains leading performance on PaNuke, despite its high inter-class visual variability.
These results highlight the superior generalization ability of our approach and its robustness across varied multi-label medical image classification tasks. By jointly learning instance features and label dependencies, our method consistently outperforms traditional MIML pipelines and demonstrates its scalability and adaptability to real-world medical applications.

\begin{table*}[htbp]
\caption{Experimental results of ablation study on the test set. The best results are in bold font. ``Granularity'' refers to the level at which sampling or learning weights are applied: \xmark \ indicates no weights (uniform sampling), $\bullet$ indicates instance-level granularity (each patch has a distinct weight),
$\circ$ indicates label-level granularity (each patch has one weight per label). \xmark, $\blacktriangle$, and $\triangle$ in the column Initialization indicate the initialization methods of weights: no weights, frequencies averaged on all data, and frequencies averaged on each sample, respectively. \xmark and \cmark in the column Sampling, Pseudo-label, and Coefficients indicate whether those components are used, respectively.}
\label{tbl:ablation_study}
\begin{center}
\begin{small}
\begin{tabular}{ccccccccccc}
\bottomrule
\textbf{Model} & Granularity & Initialization & Sampling & Pseudo-label & Coefficients & $F1_{Micro}$ & $F1_{Macro}$ & Acc. & mAP & Overall \\
\hline
1 & \xmark & \xmark & \xmark & \xmark & \xmark & 0.717 & 0.353 & 0.367 & 0.445 & 0.470 \\
\hline

2 & $\bullet$ & $\blacktriangle$ & \xmark & \xmark & \xmark & 0.550 & 0.200 & 0.299 & 0.357 & 0.352 \\

3 & $\bullet$ & $\blacktriangle$ & \xmark & \xmark & \cmark & 0.693 & 0.417 & 0.402 & 0.503 & 0.504 \\

4 & $\bullet$ & $\blacktriangle$ & \cmark & \xmark & \xmark & 0.774 & 0.613 & 0.451 & 0.692 & 0.633 \\

5 & $\bullet$ & $\blacktriangle$ & \cmark & \xmark & \cmark & 0.788 & 0.693 & 0.533 & 0.746 & 0.690 \\
\hline
6 & $\circ$ & $\blacktriangle$ & \xmark & \xmark & \xmark & 0.242 & 0.199 & 0.136 & 0.441 & 0.255 \\

7 & $\circ$ & $\blacktriangle$ & \xmark & \xmark & \cmark & 0.672 & 0.395 & 0.373 & 0.516 & 0.489 \\

8 & $\circ$ & $\blacktriangle$ & \xmark & \cmark & \xmark & 0.173 & 0.085 & 0.093 & 0.481 & 0.208 \\

9 & $\circ$ & $\blacktriangle$ & \xmark & \cmark & \cmark & 0.753 & 0.457 & 0.436 & 0.538 & 0.546 \\

10 & $\circ$ & $\blacktriangle$ & \cmark & \xmark & \xmark & 0.780 & 0.675 & 0.547 & 0.762 & 0.691 \\

11 & $\circ$ & $\blacktriangle$ & \cmark & \xmark & \cmark & 0.790 & 0.678 & 0.533 & 0.761 & 0.690 \\

12 & $\circ$ & $\blacktriangle$ & \cmark & \cmark & \xmark & 0.770 & 0.707 & 0.541 & \textbf{0.804} & 0.705 \\

13 & $\circ$ & $\blacktriangle$ & \cmark & \cmark & \cmark & 0.780 & 0.682 & 0.485 & 0.747 & 0.674 \\
\hline
14 & $\circ$ & $\triangle$ & \xmark & \xmark & \xmark & 0.321 & 0.197 & 0.175 & 0.472 & 0.291 \\

15 & $\circ$ & $\triangle$ & \xmark & \xmark & \cmark & 0.719 & 0.435 & 0.443 & 0.529 & 0.532 \\

16 & $\circ$ & $\triangle$ & \xmark & \cmark & \xmark & 0.050 & 0.084 & 0.031 & 0.464 & 0.157 \\

17 & $\circ$ & $\triangle$ & \xmark & \cmark & \cmark & 0.729 & 0.461 & 0.422 & 0.578 & 0.547 \\

18 & $\circ$ & $\triangle$ & \cmark & \xmark & \xmark & 0.757 & 0.663 & 0.543 & 0.770 & 0.683 \\

19 & $\circ$ & $\triangle$ & \cmark & \xmark & \cmark & 0.783 & 0.676 & 0.524 & 0.758 & 0.685 \\

20 & $\circ$ & $\triangle$ & \cmark & \cmark & \xmark & 0.779 & 0.685 & 0.554 & 0.778 & 0.699 \\

21 & $\circ$ & $\triangle$ & \cmark & \cmark & \cmark & \textbf{0.821} & \textbf{0.745} & \textbf{0.573} & 0.787 & \textbf{0.732} \\
\bottomrule
\end{tabular}
\end{small}
\end{center}
\end{table*}

\subsection{Ablation Study}
\label{sec:ablation}

To study the importance of different components of the proposed method, we have conducted several ablation experiments. \\

\subsubsection{Backbone Network}
\label{sec:backbone}
The backbone network is crucial for feature extraction in the proposed method. We evaluate six widely used architectures: VGG16~\cite{simonyan2015deepconvolutionalnetworkslargescale}, Inception-v3~\cite{szegedy2015rethinkinginceptionarchitecturecomputer}, ResNet-50~\cite{he2015deepresiduallearningimage}, ResNet-101~\cite{he2015deepresiduallearningimage}, CvT-w24~\cite{wu2021cvtintroducingconvolutionsvision}, and TResNet-L~\cite{ridnik2020tresnethighperformancegpudedicated}. Each backbone follows its default input resolution and is assessed under the same experimental conditions. As shown in Table~\ref{tab:backbone}, ResNet-50 achieves the best overall performance across key metrics, making it the optimal choice for integrating with our proposed approach.

\begin{table}[htbp]
    \centering
    \caption{Experimental results of different base models on the test set. The best results are in bold font.}
    \begin{tabular}{ccp{0.7cm}p{0.8cm}p{0.5cm}p{0.5cm}}
    \bottomrule
     Backbone    & Input
Resolution&$F1_{\text{Micro}}$ & $F1_{\text{Macro}}$ & Acc. & mAP  \\
    \hline
    VGG16~\cite{simonyan2015deepconvolutionalnetworkslargescale}     & $224\times224$ &0.030 &0.009 & 0.004 & 0.199\\
    Inception-v3~\cite{szegedy2015rethinkinginceptionarchitecturecomputer} & $299\times299$ &0.599 &0.185 & 0.285 & 0.323\\
    ResNet-50~\cite{he2015deepresiduallearningimage}    & $224\times224$ &\textbf{0.717} &\textbf{0.353} & \textbf{0.467} & 0.445\\
    ResNet-101~\cite{he2015deepresiduallearningimage}    & $448\times448$ &0.672 &0.311 & 0.337 & 0.421\\
    CvT-w24~\cite{wu2021cvtintroducingconvolutionsvision}    & $384\times384$ &0.698 &0.297 & 0.372 & 0.433\\
    TResNet-L~\cite{ridnik2020tresnethighperformancegpudedicated}    & $448\times448$ &0.701 &0.335 & 0.407 & \textbf{0.461}\\
    \bottomrule

    \end{tabular}
    
    \label{tab:backbone}
\end{table}

From Table \ref{tab:backbone}, it is evident that ResNet-50 achieves the best overall performance, surpassing other models in $F1_{\text{Micro}}$, $F1_{\text{Macro}}$, and Accuracy, making it the most suitable choice for ANA detection. 
TResNet-L achieves the highest mAP, indicating its strength in ranking-based evaluation, but its overall performance remains slightly behind ResNet-50. %
ResNet-101 performs competitively but does not surpass ResNet-50, likely due to increased computational complexity without a proportional gain in feature representation.

In contrast, VGG16 performs the worst across all metrics, suggesting that its shallower architecture lacks the depth required to extract meaningful features for ANA detection. 
Inception-v3 also struggles, particularly in $F1_{\text{Macro}}$, which indicates poor performance on less frequent ANA patterns. This may be attributed to its convolutional layers with a stride of 2, which enlarge the receptive field, inadvertently introducing irrelevant neighboring information and leading to misclassification.
Both CvT-w24 and TResNet-L achieve moderate performance, demonstrating that vision transformers and optimized CNN architectures have potential in ANA detection. 
However, their performance does not surpass ResNet-50, indicating that a well-balanced deep residual network remains the most effective backbone for this task.
\subsubsection{Ablation Analysis of Core Components and Weighting Strategies}
Besides the three components introduced in \ref{sec:methodlogy}, we also evaluate the influence of different initializing strategies for weights and different granularities of weights. 
There are two types of initializations for weights, namely initializations with frequencies averaged on all data and each sample, respectively. 
The former counts the frequency numbers of each ANA type over all data, and the weights of all samples share the same initial values, whereas the latter only counts that over each sample, and the weights within the same sample share the initial values. 
The formal definitions of the two initializations are written as:
\begin{equation}
\label{eq_initializations}
\begin{split}
w_k^d &= \text{softmax}(\frac{\sum_{i=1}^{N}(t_{i,k}^j)}{\sum_{k=1}^{K}\sum_{i=1}^{N}(t_{i,k}^j)}) \\
w_{p,k}^s &= \text{softmax}(\frac{\sum_{i \in S_p}(t_{i,k}^j)}{\sum_{k=1}^{K}\sum_{i \in S_p}(t_{i,k}^j)})
\end{split}
\end{equation}
where $w_k^d, w_{p,k}^s$ are initialized weights averaged on all data and each sample respectively, $\text{softmax}(\cdot)$ denotes the softmax function applied along the label dimension, and $S_p$ is the extracted patch set of $p$-th sample. Rather than weights at the label-level, weights at the instance-level, namely a scalar weight for each instance instead of a vector weight, are also evaluated in the ablation study. Due to the enlarged granularity, the previous initialization of weights $w_k^d$ is not applicable and changed to:

\begin{equation}
\label{eq_granularity}
w_{q,k}^d = \text{softmax}(\frac{N_{L_q}}{N})
\end{equation}
where $w_{q,d}^d$ denotes the new initialized weights averaged on all data at the instance-level and $N_{L_q}$ is the number of instances, which share the same label $L_q \in \mathbb{Z}^K$, where $K$ represents the total number of categories of labels. Besides, in the case of weights at the instance level, it is not able to implement the initialization of frequencies averaged on each sample and the probabilistic pseudo-label dispatcher, the corresponding experimental results of which are not reported in this section.

From Table \ref{tbl:ablation_study}, it is evident that the proposed components effectively improve the performance of the model. From the perspective of granularities of weights, all experiments in the ablation study except for the first one can be divided into two groups, one is the case of weights at the instance-level and the other is that at the label-level ($\bullet$ and $\circ$ in column Granularity). For the former group, although the performance is limited due to the granularity of weights, the effectiveness of the instance sampler and label-aware coefficients can be observed. For the latter group, the experiments are further organized into two subgroups ($\blacktriangle$ and $\triangle$ in column Initialization) in terms of the initialization method of weights formulated in Eq. \ref{eq_initializations}. Only a slight performance difference between the two initialization methods can be seen from the two subgroup experiments, where significant performance improvement has been demonstrated for the proposed three components. The instance sampler contributes most to the performance improvement, while the probabilistic pseudo-label and label-aware coefficients usually provide minor improvement. Among all experiments, the instance sampler improves the overall performance by at least 23.44\%. The probabilistic pseudo-label contributes positively when models are stable, which ensures the correctness of generated pseudo-labels and avoids wrong leads. Besides improving the performance, label-aware coefficients of weight learning rates also effectively stabilize the learning procedure and prevent it from being stuck in local optima.

\subsubsection{Effectiveness of Instance Sampler}
\begin{figure}[htbp]
\centering
\includegraphics[width=0.5\textwidth]{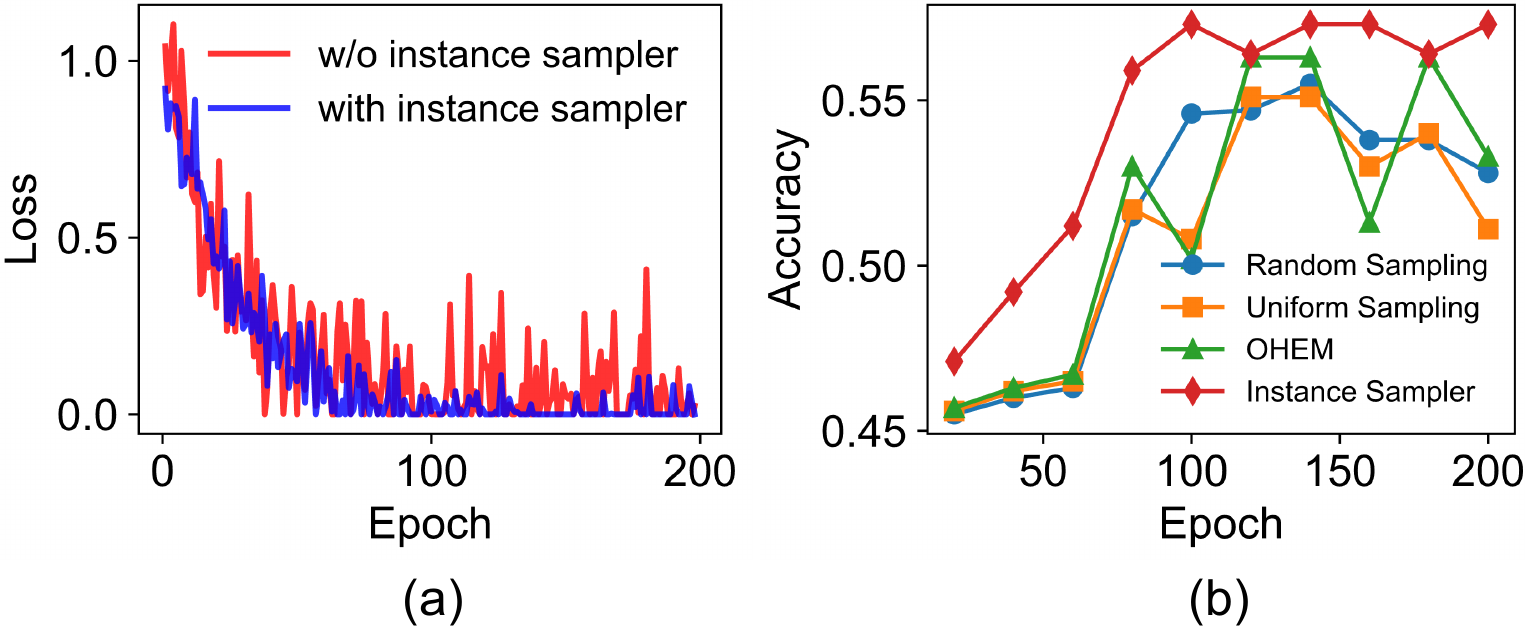}
\caption{(a) Training loss curves comparing the model with and without the instance sampler. (b) Performance results under different sampling strategies: Random, Uniform, and OHEM. }
\label{fig:sampleeg}
\end{figure}

In Fig.~\ref{fig:sampleeg} (a), we present the training loss curves comparing the model with and without the instance sampler, showing that our sampler not only accelerates convergence but also stabilizes the training process. 
Meanwhile, Fig.~\ref{fig:sampleeg} (b) reports the final accuracy under various sampling strategies, including random sampling, uniform sampling, and OHEM~\cite{shrivastava2016trainingregionbasedobjectdetectors}. The results reveal that our instance sampler consistently outperforms these alternatives in the multi-label ANA classification task.

Our instance sampler achieves higher overall performance than both the Random and Uniform baselines, indicating that merely relying on random or evenly distributed sampling fails to exploit the inherent diversity among subregions. 
By contrast, our method effectively identifies and prioritizes subregions that exhibit representative ANA patterns. 
Although OHEM does focus on high-loss samples to some extent, it relies primarily on loss value and overlooks the interplay or co-occurrence between different ANA types in a multi-label setting. 
In comparison, our instance sampler leverages self-paced learning and the dynamic weighting of $\alpha_i^j$, enabling a more comprehensive assessment of subregion value in multi-label tasks and thereby yielding superior results.\\

\subsubsection{Sensitivity Analysis of Label-aware Scaling Parameter $C$}
\begin{table}[t]
\centering
\caption{Performance comparison under different values of $C$.}
\label{tab:diff_C}
\begin{tabular}{c|cc}
\toprule
\textbf{$C$} & \textbf{Accuracy (\%)} & \textbf{mAP (\%)} \\
\midrule
2 & 51.2 & 74.5 \\
3 & 52.4 & 75.6 \\
4 & 54.1 & 76.2 \\
5 & 55.5 & 76.9 \\
6 & \textbf{57.3} & \textbf{78.7} \\
7 & 45.1 & 62.5\\
\bottomrule
\end{tabular}
\end{table}

We also explore the impact of different values of $C$ on model performance, as shown in Table~\ref{tab:diff_C}. Since $C$ determines the maximum number of coexisting labels per sample, it directly influences the label-aware coefficient $\lambda$, thereby adjusting the learning rate of weights. 
Intuitively, increasing $C$ allows the model to better handle multi-label complexity by adapting to more challenging cases with higher learning rates. Gradually increasing $C$ from 2 to 6 consistently improves both accuracy and mAP, suggesting that enabling the model to learn from more complex multi-label samples enhances its generalization ability. 
Notably, performance saturates and peaks at $C=6$, which aligns with the actual maximum number of coexisting labels in our dataset. However, when $C$ is set to 7, a significant performance drop is observed, likely due to excessive learning rates leading to instability. 
This finding highlights the importance of setting an appropriate $C$ value to balance learning effectiveness and model stability.\\

\subsubsection{Sensitivity to Patch Size}
To investigate the effect of patch size on classification performance, we evaluate our method using patch resolutions ranging from $32\times32$ to $640\times640$, along with the original unpatched image. 
All patches are resized to $224\times224$ before being fed into the ResNet backbone. As shown in Table~\ref{tab:sen}, extremely small patches lead to massive instance counts and degraded performance due to excessive noise. 
Medium-sized patches (e.g., $224\times224$ and $448\times448$) offer a better balance of localization and context, achieving the best results across metrics. In particular, $448\times448$ yields the highest $F1_{\text{Macro}}$ and mAP. 
Additionally, we include the average number of subregions ($N$) per image to account for variation in image resolution, as different patch sizes produce different numbers of instances.

\begin{table}[htbp]
    \centering
    \caption{Sensitivity analysis of different patch sizes}
    \begin{tabular}{>{\centering\arraybackslash}p{1.3cm}>{\centering\arraybackslash}p{1.8cm}>{\centering\arraybackslash}p{0.5cm}p{0.7cm}p{0.7cm}p{0.5cm}p{0.5cm}}
    \bottomrule
    \rule{0pt}{8pt}
    Patch Size & Input instances & $\bar{N}$ & $F1_{\text{Micro}}$ & $F1_{\text{Macro}}$ & Acc. & mAP  \\
    \hline
    $32\times32$ &22,158,220  &4811& 0.080 &0.012 & 0.011 & 0.179\\
     $64\times64$ & 5,597,220  & 1215& 0.088 &0.025 & 0.017 & 0.199\\
     $128\times128$ & 1,390,620  & 301&0.440 &0.215 & 0.277 & 0.332\\
     $224\times224$ &412,250  &89&\textbf{0.840} &0.725 & 0.554 & 0.760\\
     $448\times448$ & 82,240  &17&0.821 & {\textbf{0.745}} & {\textbf{0.573}} & \textbf{0.787}\\
     $640\times640$ & 40,985  &8&0.792 &0.729 & 0.522 & 0.701\\
     $\text{Original}$ & 4,605  &8&0.672 &0.311 & 0.337 & 0.421\\
    \bottomrule
    \end{tabular}
    \label{tab:sen}
\end{table}

\begin{figure}[htbp]
    \centering
    \includegraphics[width=\linewidth]{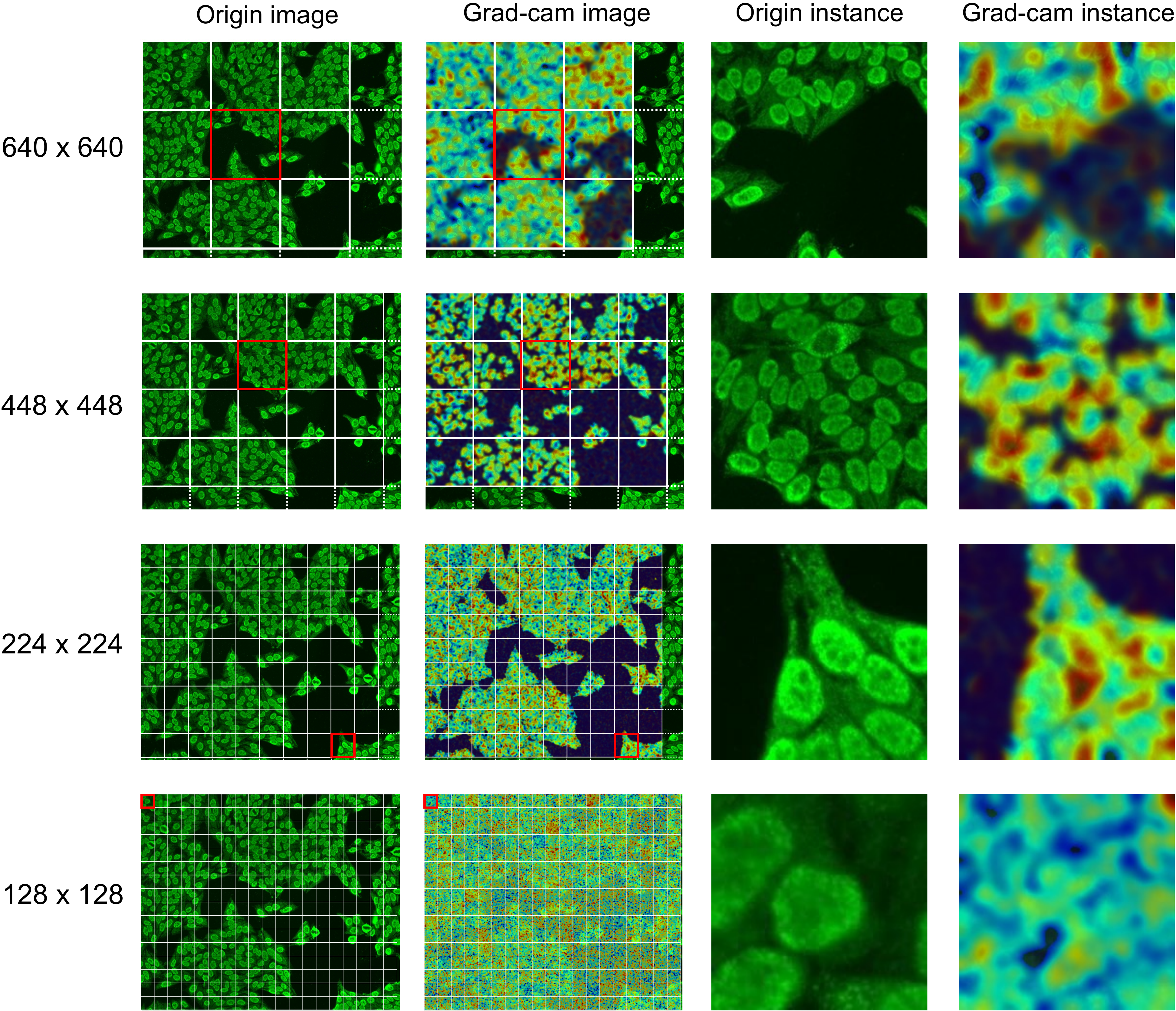}
    \caption{
        \textbf{Grad-CAM visualization for different patch sizes.} Patch size of $448\times448$ demonstrates the most focused activation on foreground ANA patterns.
    }
    \label{fig:gradcam}
\end{figure}

To provide further insight, we visualize Grad-CAM~\cite{selvaraju2017grad} heatmaps under different patch sizes in Figure~\ref{fig:gradcam}. The results reveal that $448\times448$ best localizes class-discriminative regions, validating its effectiveness in capturing informative foreground content.

\section{Conclusion}
In this paper, we have proposed a new framework consisting of three novel modules to address the multi-instance multi-label challenge of ANA detection.
Empirical results on the ANA dataset demonstrate that the proposed framework consistently outperforms existing state-of-the-art approaches across a range of metrics and settings.
The comparison results on other MIML medical image domains verify the generalization ability of our proposed framework. 
Despite these promising results, the current framework relies on heuristic-based confidence estimation and fixed-size patch sampling, which may overlook nuanced patterns in more complex visual contexts. 
In the future, we plan to explore continuous attention mechanisms to better capture nuanced visual patterns and improve the understanding of how pseudo-labels are assigned. We will also extend our framework to more medical imaging tasks.

\bibliographystyle{IEEEtran}
\bibliography{IEEEabrv,refs}

\end{document}